\documentclass[10pt,twocolumn,letterpaper]{article}

\usepackage[pagenumbers]{wacv} 
\usepackage{multirow}
\usepackage{amsmath}
\usepackage{pifont}
\definecolor{wacvblue}{rgb}{0.21,0.49,0.74}
\usepackage[pagebackref,breaklinks,colorlinks,allcolors=wacvblue]{hyperref}

\def\wacvPaperID{2201} 
\def\confName{WACV}
\def\confYear{2026}

\newcommand{\cmark}{\ding{51}}%
\newcommand{\xmark}{\ding{55}}%
\title{PRNU-Bench: A Novel Benchmark and Model for PRNU-Based Camera Identification}

\author{Florinel Alin Croitoru, Vlad Hondru, Radu Tudor Ionescu\\
University of Bucharest, Romania\\
Corresponding author: {\tt raducu.ionescu@gmail.com}
}

\begin{document}
\maketitle

\begin{abstract}
We propose a novel benchmark for camera identification via Photo Response Non-Uniformity (PRNU) estimation. The benchmark comprises 13K photos taken with 120+ cameras, where the training and test photos are taken in different scenarios, enabling ``in-the-wild'' evaluation. In addition, we propose a novel PRNU-based camera identification model that employs a hybrid architecture, comprising a denoising autoencoder to estimate the PRNU signal and a convolutional network that can perform 1:N verification of camera devices. Instead of using a conventional approach based on contrastive learning, our method takes the Hadamard product between reference and query PRNU signals as input. This novel design leads to significantly better results compared with state-of-the-art models based on denoising autoencoders and contrastive learning. We release our dataset and code at: 
\url{https://github.com/CroitoruAlin/PRNU-Bench}.
\end{abstract}

\setlength{\abovedisplayskip}{2.0pt}
\setlength{\belowdisplayskip}{2.0pt}

\section{Introduction}
\label{sec:intro}
Device usage with digital imaging capabilities has been growing exponentially, and with the ubiquitous nature of smartphones, the ways in which visual content is created, shared, and consumed have expanded. Moreover, image manipulation techniques have rapidly evolved, with the generative AI playing an important role \cite{Croitoru-TPAMI-2023, karras-CVPR-2019, karras-CVPR-2020, karras-NeurIPS-2021, mou-CVPR-2024, shi-CVPR-2024, huang-TPAMI-2025, feng-AAAI-2025}. This digital revolution raised concerns about media authenticity\footnote{\url{https://c2pa.org/}}
and provenance \cite{rossler-ICCV-2019, sareen-CRC-2022, feng-ACN-2023}. Hence, the ability to reliably determine the source camera of a digital image has become a critical asset in fighting against counterfeiting in various domains, \eg~journalism.

Photo-Response Non-Uniformity (PRNU) has become one of the most reliable device-specific fingerprint in multimedia forensics, and thus, it was widely adopted in techniques for camera identification and image authentication~\cite{lukas-TIFS-2006,Akbari-ICPR-2022, cozzolino-JIS-2020,Cozzolino-TIFS-2018, Kirchner-WIFS-2019}. The method exploits the unique multiplicative noise pattern of each digital camera sensor, a consequence of imperfect manufacturing~\cite{lukas-TIFS-2006}.
Given that PRNU is specific to each device, in order to develop reliable solutions for camera identification, datasets~\cite{Thomas-SAC-2010, Bernacki-SECRYPT-2023, Bruno-MTA-2023, Hadwiger-ICPR-2021, Shullani-IS-2017} have to be continuously updated with the latest devices, as well as include a wide range of sensors. Yet, recent datasets~\cite{Bernacki-SECRYPT-2023, Bruno-MTA-2023, Hadwiger-ICPR-2021} comprise relatively outdated devices, the newest device from these datasets being manufactured in 2020. Another common issue of existing datasets is the focus on specific camera types. For instance, UNISA20~\cite{Bruno-MTA-2023} or FODB~\cite{Hadwiger-ICPR-2021} contain multiple instances of a single model or device type. This limits the ability to evaluate how well methods generalize across different camera models and device types. Moreover, existing datasets~\cite{Thomas-SAC-2010, Bernacki-SECRYPT-2023, Bruno-MTA-2023, Hadwiger-ICPR-2021, Shullani-IS-2017} do not ensure consistent collection setups across devices, which is crucial to prevent methods from exploiting content differences rather than device characteristics. To combat these issues, we propose \textbf{PRNU-Bench}, a novel benchmark for source camera identification. Our dataset includes a wider and newer range of 126 cameras (some released in 2024), and its collection protocol and data separation are explicitly designed to mitigate spurious correlations \cite{Izmailov-NeurIPS-2022}.

\begin{figure}[t]
    \centering
    \includegraphics[width=1.\linewidth]{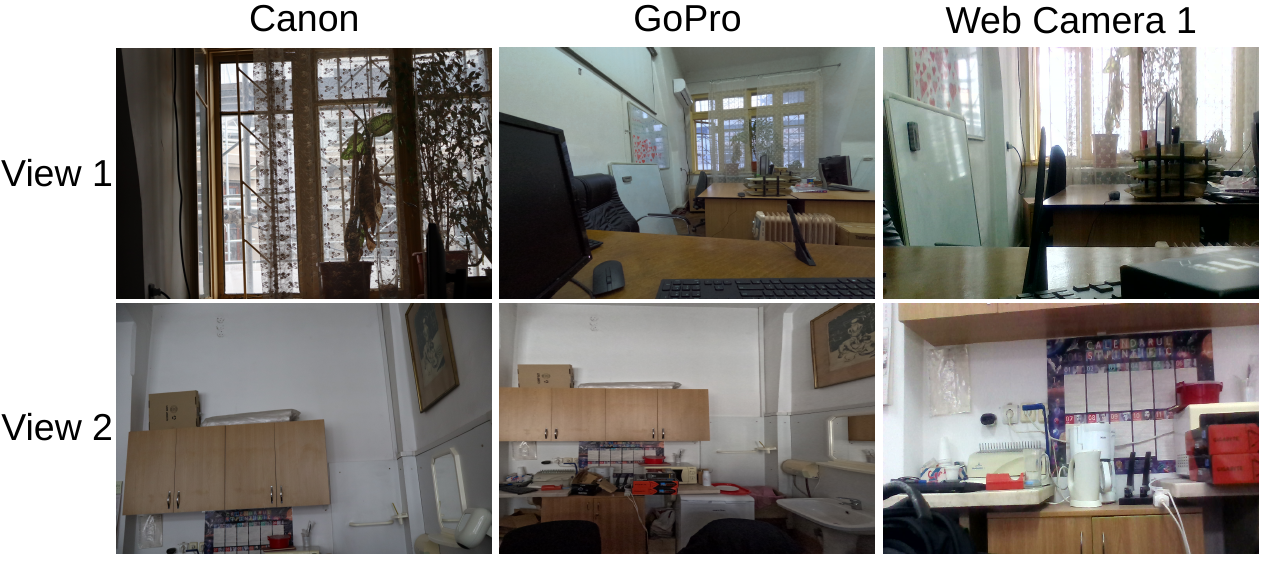}
    \vspace{-0.7cm}
    \caption{Example images from PRNU-Bench. Each row corresponds to a different side of the office where photos were taken, while each column shows images captured by the same sensor.}
    \label{fig:samples}
    \vspace{-0.3cm}
\end{figure}

\begin{table*}[t]
  \centering
  \small{
  \begin{tabular}{ccccccccc}
    \toprule
    \multirow{2}{*}{ Dataset} & Sensor  & Image & \multicolumn{2}{c}{Resolution} & Flat & Distinctive & \multicolumn{2}{c}{Device}\\
    & Count & Count & Min & Max & Images & Scenes & Oldest & Newest \\
    \midrule
    Dresden~\cite{Thomas-SAC-2010} & $73$ & $15,025$ & $2592 \times 1944$ & $4032 \times 3024$ & No & Yes & 2004 & 2009 \\
    Imagine~\cite{Bernacki-SECRYPT-2023} & $55$ & $2,500$ & $2560 \times 1920$ & $8640 \times 5760$ & No & No & 2013 & 2020 \\
    UNISA2020~\cite{Bruno-MTA-2023} & $20$ & $4,647$ & $4288 \times 2848$ & $4288 \times 2848$ & No & Yes & 2010 & 2010 \\
    FODB~\cite{Hadwiger-ICPR-2021} & $27$ & $23,000$ & $2048\times1536$ & $4000\times3000$ & No & Yes & 2008 & 2019\\
    VISION~\cite{Shullani-IS-2017} & $35$ & $34,427$ & $960 \times 720$ & $5312   \times   2988$ & Yes & Yes & 2010 & 2016\\
    \midrule
    PRNU-Bench (Ours) & $126$ & $12,960$ & $1920 \times 1080$ & $5568 \times 4872$ & No & Yes & 2016 & 2024 \\
    \bottomrule
  \end{tabular}
  }
  \vspace{-0.25cm}
  \caption{Comparison of PRNU-Bench with other datasets for source camera identification. \textit{Flat images} are marked true when the subset used for PRNU estimation contains flat images, which are not practical for real-world scenarios. \textit{Distinctive scenes} are marked true when the PRNU registration and identification sets consist of disjoint scenes.}
  \label{tab:comparison}
  \vspace{-0.2cm}
\end{table*}

Camera identification leveraging PRNU was originally proposed by Lukas \etal~\cite{lukas-TIFS-2006} in 2006, demonstrating the efficiency of exploiting these manufacturing imperfections of the image sensors. Since then, scientists have made sustained efforts in this domain to evolve the models from a research concept to a forensic tool in practical applications \cite{chen-TIFS-2008, Goljan-DW-2009, Mehrish-SPL-2016, Mehrish-SPIC-2019, Marra-WIFS-2016, Marra-TIFS-2017, Kirchner-WIFS-2019, Zeng-arXiv-2021, Cozzolino-TIFS-2018, cozzolino-JIS-2020, nayerifard-JS-2025, Akbari-ICPR-2022, Benegui-ACNS-2020}. Nevertheless, these methods aim to improve the PRNU fingerprint extraction, typically by proposing deep learning architectures and novel training strategies~\cite{chen-TIFS-2008, Cozzolino-TIFS-2018, cozzolino-JIS-2020, Kirchner-WIFS-2019}. In contrast, our work does not focus on improving the extraction process itself, but rather on how the extracted PRNU fingerprint is leveraged. As such, our contribution is orthogonal to prior advances in fingerprint extraction. Recently, Nayerifard~\etal~\cite{nayerifard-JS-2025} presented a PRNU-based classifier with a similar objective, improving the use of the extracted fingerprint. However, because their approach relies on classification, it is restricted to devices seen during training. In contrast, our method generalizes to previously unseen cameras by identifying matching pairs of PRNU fingerprint and noise residuals by analyzing their Hadamard product.

We carry out experiments on PRNU-Bench (ours) and Dresden \cite{Thomas-SAC-2010} datasets to compare the proposed method with a range of baselines \cite{Goljan-EI-2009,Zamir-CVPR-2022,Zhang-TPAMI-2021}. The empirical results attest the superiority of our approach over existing methods. Moreover, we conduct ablation studies to assess the contributions of individual components. The ablation results justify the proposed design. 



In summary, our contribution is threefold:
\begin{itemize}
    \item We introduce PRNU-Bench, a dataset containing images from a wide range of new devices, collected through a process specifically designed to support a realistic and challenging evaluation.
    \item We propose a novel model that jointly analyzes pixel-level correlations via the Hadamard product between PRNU fingerprints and noise residuals.
    \item We demonstrate significant performance improvements compared with state-of-the-art methods, raising the top-1 accuracy by roughly $14\%$ on PRNU-Bench. We also show that our method generalizes well, as confirmed by its strong performance on the Dresden dataset.
\end{itemize}

\section{Related Work}
Photo Response Non-Uniformity represents the unique noise pattern of an individual digital camera sensor, acting as a device-specific fingerprint. This arises from manufacturing imperfections in individual photosites. Using PRNU to identify the source camera was first demonstrated by Lukas \etal~\cite{lukas-TIFS-2006}, and then improved by the subsequent works of Chen \etal~\cite{chen-TIFS-2008} and Golijan \etal~\cite{Goljan-DW-2009}. These early contributions established PRNU estimation as a cornerstone of multimedia forensics.

\noindent
\textbf{PRNU Datasets.} PRNU research has a long history in digital image forensics. Yet, the development of dedicated datasets remains an ongoing effort \cite{Shullani-IS-2017, Thomas-SAC-2010, Bernacki-SECRYPT-2023, Bruno-MTA-2023, Hadwiger-ICPR-2021}, as new collections continue to emerge in order to keep benchmarks relevant, reflecting the characteristics of the latest imaging devices and processing pipelines.
One of the earliest and most widely used datasets for PRNU research is Dresden~\cite{Thomas-SAC-2010}. Although it consists of many different sensors from a wide range of device types, these devices are beyond their end-of-life cycle. The Imagine dataset~\cite{Bernacki-SECRYPT-2023} tries to address this limitation, seeking to include more modern devices, such as smartphones, compact cameras or DSRLs. Given that recent PRNU estimation methods are based on data-driven and machine learning approaches, datasets had to be accommodated as well. For example, VISION~\cite{Shullani-IS-2017} introduced PRNU evaluation in videos, where compression and stabilization pose extra challenges. UNISA2020~\cite{Bruno-MTA-2023} focuses on scenarios with different instances of the same camera model, although its devices are relatively old. Finally, FODB~\cite{Hadwiger-ICPR-2021} comprises images of different scenes, enabling a realistic evaluation, as well as including post-processed images, focusing on operations applied to the photos shared on social media applications.


Although various datasets are publicly available~\cite{Shullani-IS-2017, Thomas-SAC-2010, Bernacki-SECRYPT-2023, Bruno-MTA-2023, Hadwiger-ICPR-2021}, there is no up-to-date benchmark to evaluate PRNU for new devices (since 2021), that appear on a yearly basis. Furthermore, some of the existing datasets~\cite{Hadwiger-ICPR-2021, Bruno-MTA-2023} lack a wide range of devices, thus having a limited scope, which makes the evaluation not robust enough. In contrast, PRNU-Bench comprises a high number of sensors from newer devices with higher resolutions, as shown in Table~\ref{tab:comparison}. Unlike VISION~\cite{Shullani-IS-2017}, PRNU-Bench does not contain the so-called ``flat images'', which are uniform white images from which the PRNU can be more easily estimated. Instead, our dataset contains only real-scene images to ensure a realistic evaluation setup. Furthermore, PRNU-Bench employs a consistent data collection process across devices, resulting in images with similar content. This design choice ensures that source camera identification methods are strongly penalized if they rely on content-based cues rather than device-specific fingerprints.

 

\noindent
\textbf{Source camera identification.} The seminal work in PRNU estimation \cite{lukas-TIFS-2006} not only introduced the idea of the noise pattern inherited by every imaging sensor, but also presented a method to determine the origin of the camera. It consists of applying a denoising filter to multiple images to suppress scene content, and thus estimate the reference noise pattern. The reference PRNU can then be correlated with the noise residuals of query images. Based on a similar idea, Chen \etal~\cite{chen-TIFS-2008} formalized a framework that integrates the statistical properties of PRNU, showing better results while also requiring fewer input images. Some limitations of this framework were addressed by Golijan \etal~\cite{Goljan-DW-2009}, who introduced \textit{false acceptance probability estimation}, making results quantifiable and usable in real forensic settings. The theory of the former approaches is based on Maximum Likelihood Estimation (MLE). Several subsequent studies employed MLE as well \cite{Mehrish-SPL-2016, Mehrish-SPIC-2019, Marra-WIFS-2016, Marra-TIFS-2017}.

More recent studies adopted deep learning  approaches \cite{Kirchner-WIFS-2019, Zeng-arXiv-2021, Cozzolino-TIFS-2018, cozzolino-JIS-2020, nayerifard-JS-2025, Akbari-ICPR-2022} to extract the noise pattern. Kirchner \etal~\cite{Kirchner-WIFS-2019} proposed to estimate the noise directly by a convolutional neural network using the noise residuals extracted with an MLE approach as target. Cozzolino \etal~\cite{cozzolino-JIS-2020} argued that PRNU and noiseprint are complementary, thus leveraging both input sources in a Siamese architecture based on CNNs. Akbari \etal~\cite{Akbari-ICPR-2022} introduced a PRNU-based layer that can be integrated into a CNN. The noise residuals are first extracted from multiple consecutive video frames with a wavelet-based denoising filter and then fed into the network, thus incorporating temporal information. Similarly, Nayerifard \etal~\cite{nayerifard-JS-2025} supplied the sensor noise as input to a CNN, but the final prediction also integrates a classical correlation-based PRNU matching. 

Source camera identification involves two main stages, reference PRNU extraction (registration) and comparison with a test image (identification). While most prior studies~\cite{Cozzolino-TIFS-2018, Akbari-ICPR-2022, Kirchner-WIFS-2019} focused on improving extraction, our method strengthens the comparison stage, making prior advances in PRNU extraction orthogonal to ours. The closest approaches to our own are the classifier of Nayerifard \etal~\cite{nayerifard-JS-2025} and the method of Cozzolino \etal~\cite{cozzolino-JIS-2020}, which leverage noiseprints linked to camera models. Unlike Nayerifard \etal~\cite{nayerifard-JS-2025}, we exploit the Hadamard product between fingerprints and noise residuals, training the model to recognize matches by analyzing their Hadamard product. Thus, our method generalizes to unseen devices without retraining. Similar to Cozzolino \etal~\cite{cozzolino-JIS-2020}, we leverage hierarchical information when computing the final similarity score between the fingerprint and the noise residuals. However, we do not rely on additional camera-specific features. Instead, our approach integrates multi-resolution representations of the PRNU signal to improve identification performance. 

\begin{table}[t]
  \centering
  \small{
  \begin{tabular}{lll}
    \toprule
    Device Model  & Type & Resolution \\
    \midrule
    Trust Tyro Full HD & Webcam & $1920 \times 1080$ \\
    Hama C-400 & Webcam & $1920 \times 1080$ \\
    Canon EOS R100 & Mirrorless& $6000 \times 4000$\\
    GoPro Hero Black 11 & Video Camera & $5568 \times 4872$\\
    Google Pixel 9 & Smartphone & $3072 \times 4080$ \\
    iPad Pro 4th Gen & Tablet & $3024 \times 4032$ \\
    iPad Pro 6th Gen & Tablet & $3024 \times 4032$ \\
    iPhone 12 & Smartphone & $3024 \times 4032$ \\
    iPhone 13 Pro & Smartphone & $3024 \times 4032$ \\
    iPhone 14 Pro & Smartphone & $3024 \times 4032$ \\
    iPhone 15 & Smartphone & $4284 \times 5712$ \\
    iPhone 6 Plus & Smartphone &  $2448 \times 3268$\\
    iPhone SE & Smartphone & $3024 \times 4032$ \\
    Nokia G21 & Smartphone & $1836 \times 4080$ \\
    OnePlus Nord & Smartphone & $3000 \times 4000$ \\
    Redmi 12C & Smartphone & $3072 \times 4080$ \\
    Samsung S10 Edge+ & Smartphone & $2184 \times 4608$\\
    Samsung Note 10+ & Smartphone & $2184 \times 4608$\\
    Samsung S6 Edge+ & Smartphone & $2988 \times 5132$\\
    \bottomrule
  \end{tabular}
  }
  \vspace{-0.25cm}
  \caption{Device models and specific image resolutions in PRNU-Bench.}
  \label{tab:devices}
  \vspace{-0.2cm}
\end{table}


\section{Dataset}
\label{dataset}
\noindent
\textbf{Overview.} We introduce PRNU-Bench, a dataset comprising 12,960 images collected from 126 unique sensors from 114 devices, hence including both single-camera and multi-camera devices. For each sensor, we collected approximately 100 images in an office environment: 50 captured from one side of the office and 50 from the opposite side. This split guarantees that the two subsets contain entirely distinct content.
Furthermore, we do not include flat images (as in the VISION~\cite{Shullani-IS-2017} dataset) in our dataset, \ie PRNU-Bench only includes natural scene images. Example images from PRNU-Bench are shown in Figure~\ref{fig:samples}. Overall, PRNU-Bench has three key properties which make it both challenging and suitable for real-world evaluation. First, the data split based distinct views for registration and identification guarantees that methods exploiting visual cues from the image content are penalized, since such cues are irrelevant across distinct views. Second, by collecting scenes from the same environment with all devices, we prevent spurious correlations linked to device-oriented image distributions. For instance, we avoid scenarios where pictures taken with one device are from a vacation in Rome, and pictures taken with another are from an office building in New York, which could allow methods to rely on scene type rather than device fingerprint. Finally, as previously noted, the images used for PRNU fingerprint extraction depict natural scenes, aligning with practical real-world conditions.

PRNU-Bench covers the widest range of sensors, and is the only dataset that includes devices released between 2021 and 2024 (see Table~\ref{tab:comparison}). It also offers greater device diversity than recent alternatives, such as FODB~\cite{Hadwiger-ICPR-2021} and UNISA2020~\cite{Bruno-MTA-2023}. While UNISA2020 contains images from a single camera model and FODB is limited to smartphones, PRNU-Bench spans multiple categories, including smartphones, tablets, webcams, video cameras, and mirrorless cameras (see Table~\ref{tab:devices}). This broader device range, together with the inclusion of newer device models, introduces greater variability in both image resolution and image quality. 


\noindent
\textbf{Data split.} As hinted earlier, we split the PRNU-Bench dataset based on the depicted views, where each view corresponds to a different side of the office where images are captured. For PRNU fingerprint extraction (during device registration), we only use images from the first view, reserving those from the second view exclusively as test samples (during device identification).
The first view subset contains 6,430 images, while the second contains 6,530.
\begin{figure*}[t]
    \centering
    \includegraphics[width=0.895\linewidth]{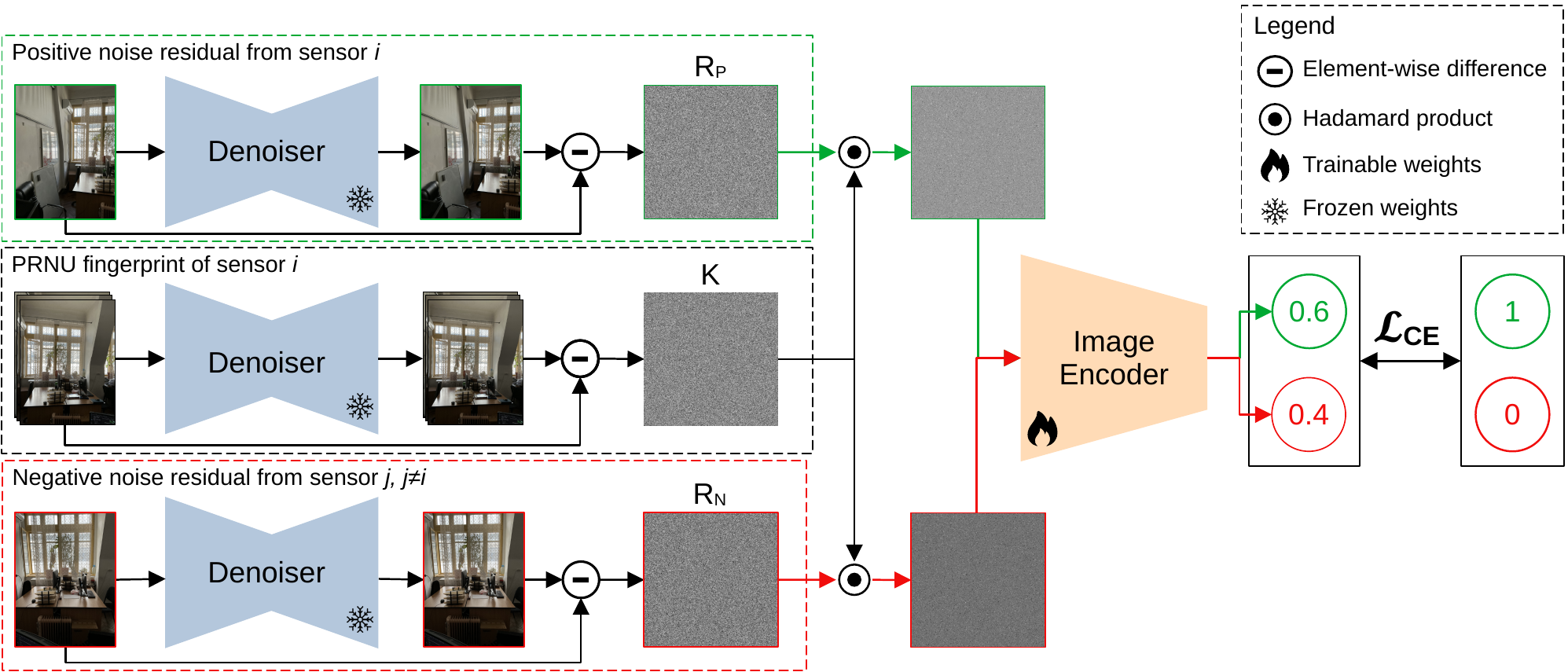}
    \vspace{-0.25cm}
    \caption{Our training pipeline. We estimate the PRNU fingerprint $K$ for a given sensor $i$. From an image captured by the same sensor $i$, we extract a positive noise residual $R_P$, while from an image captured by a different sensor $j \neq i$, we extract a negative noise residual $R_N$. A neural classifier is then trained to discriminate between the Hadamard products $K \odot R_P$ and $K \odot R_N$.}
    \label{fig:method}
    \vspace{-0.4cm}
\end{figure*}

In addition to the view-based split, we define separate pre-training and evaluation partitions. This split is crucial for neural-based methods, which require a pre-training phase before the model can compare fingerprints with noise residuals. To properly assess generalization to unseen devices, the evaluation set must contain entirely different devices from those used during pre-training, as illustrated in Figure~\ref{dataset}. 
Accordingly, we randomly assign 50 sensors to the pre-training set and reserve the remaining 76 sensors exclusively for evaluation. Since the second view is dedicated to device identification, the corresponding images are excluded from pre-training, leaving only the first view for model pre-training. This setup prevents overfitting the hyperparameters to the test view. For each of the 50 devices, we use a small subset of first-view images to extract the reference PRNU fingerprint, while the remaining first-view images serve as pre-training samples. In contrast, the evaluation partition includes both first-view and second-view images. As noted earlier, first-view images are used for reference PRNU fingerprint extraction, while noise residuals extracted from second-view images serve as test samples. To reflect realistic scenarios where only a few images are available for reference fingerprint estimation, we limit this step to $5$ images per device in our experiments, for both pre-training and evaluation.


\section{Method}
\label{method}
\noindent
\textbf{Preliminaries and basic approach.} The PRNU signal of a sensor is generally estimated from a collection of $n$ images $\{I_i\}_{i=1}^n$ acquired by that sensor. For each image, we first compute its residual noise $R_i$, which captures high-frequency components dominated by sensor-specific artifacts. This is achieved by subtracting from the original image $I_i$ its denoised counterpart $D_\theta(I_i)$, where $D_\theta$ denotes a denoising network parameterized by weights $\theta$:
\begin{equation}
\label{residual_noise}
R_i = I_i - D_\theta(I_i), \quad \forall i \in \{1, \dots, n\}.
\end{equation}
Intuitively, this operation suppresses the scene content while preserving the sensor noise. The estimated PRNU fingerprint $K$ is then obtained by averaging the residuals across all $n$ images:
\begin{equation}
\label{prnu_estimate}
K = \frac{1}{n}\sum_{i=1}^n R_i.
\end{equation}
This aggregation step reduces the influence of random noise while reinforcing the device-specific PRNU pattern. In practice, $K$ can undergo additional post-processed steps. Following previous work~\cite{chen-TIFS-2008}, we apply Wiener filtering to post-process $K$.

To determine whether a new image $A$ originates from the sensor characterized by fingerprint $K$, the normalized cross-correlation (NCC) is  usually employed. First, we extract the residual noise $R_A$ from $A$ by applying Eq.~\eqref{residual_noise}. Then, we evaluate the similarity between $R_A$ and $K$ by computing the NCC, as defined below:
\begin{equation}
    \label{normalized cross-correlation}
    \text{NCC}(K, R_A) = \frac{\left\langle \left(K-\bar{K}\right), \left(R-\bar{R}_{A}\right) \right\rangle}{\left\Vert K-\bar{K}\right\Vert \cdot \left\Vert R-\bar{R}_{A}\right\Vert},
\end{equation}
where $\langle\cdot,\cdot\rangle$ denotes the dot product, and $\bar{K}$ and $\bar{R}_{A}$ represent the pixel-wise means of $K$ and $R_A$, respectively.

\noindent
\textbf{Neural-based comparison.} For optimal accuracy, the comparison based on NCC typically requires more than 50 reference images for camera fingerprint estimation, as reported in \cite{lukas-TIFS-2006}, which can be impractical in some real-world cases. In this work, our objective is to improve the previous NCC-based comparison of raw signals with an approach based on neural networks. The motivation behind this change is to enhance the matching performance, especially in scenarios where the fingerprint $K$ is estimated from a limited number of images ($n$ is small). 

As in face recognition, a natural solution to build a neural network that can generalize to new identities is to employ contrastive learning \cite{Deng-CVPR-2019,Luan-TIP-2023,Schroff-CVPR-2015}. However, unlike faces, PRNU fingerprints are generally weak signals, and they vary a lot from one device to another. This makes it difficult to extract generalizable patterns from PRNU signal tuples via contrastive learning. Instead of comparing PRNU signals, we propose to look at pixel-level correlations between the compared signals, which can be computed via the Hadamard product between PRNU signals. By analyzing the pixel-level correlations of both correlated and uncorrelated PRNU signal pairs, we conjecture that the neural network has to solve a much easier task: to weight and integrate the pixel-level correlations in order to classify PRNU signal pairs as belonging to the same device or not. We provide empirical evidence to support our conjecture, which motivates the proposed design.

Given a set of $d$ devices with corresponding fingerprints $\{K_i\}_{i=1}^d$, our solution to the aforementioned comparison is a binary classifier that evaluates the Hadamard product between camera fingerprints and noise residuals. For a training batch, we randomly sample a fingerprint $K_i$, with $i \sim \mathcal{U}(\{1,\dots,d\})$. As a positive sample $P$, we select an image captured by the same device (with index $i$). As a negative sample $N$, we take an image from a randomly chosen device $j$, such that $j \neq i$. The corresponding residuals $R_P$ and $R_N$ are then computed using Eq.~\eqref{residual_noise}. We construct the inputs $K_i \odot R_P$ and $K_i \odot R_N$ and feed them into a vision encoder, assigning label $1$ to the former and $0$ to the latter. In this setup, we perform training via the binary cross-entropy loss, as defined below:
\begin{equation}
\label{binary_CE}
  \begin{split}
    \mathcal{L}_{\text{CE}}(K_i, R_P, R_N) =& - \log\left(E_\phi(K_i \odot R_P)\right)\\ & - 
    \log\left(1-E_\phi(K_i \odot R_N)\right),
\end{split}    
\end{equation}
where $\odot$ denotes the Hadamard product, and $E_\phi$ is the vision encoder parameterized by $\phi$. The loss defined in Eq.~\eqref{binary_CE} optimizes the learnable parameters $\phi$ to discriminate between the two products of PRNU signal pairs. In Figure~\ref{fig:method}, we illustrate the entire training pipeline, including how samples are formed to train the image encoder. At inference time, we input the Hadamard product between a camera fingerprint and the noise residual of the test image into $E_\phi$ and we use the output as a similarity score to decide if there is a match between the two or not.

\noindent
\textbf{Multi-resolution.} The PRNU signal is a fixed sensor-specific spatial pattern that is sensitive to image resizing. Indeed, resizing the image via interpolation can distort or partially suppress the fingerprint. The resizing operation acts like a low pass filtering, so different resizing scales tend to preserve different components of the signal. Noise signals that resist image downsampling are less invariant to image transformations. Moreover, we empirically observed that PRNU-based identification is still possible when using downsampled images, although the performance significantly degrades. Since downsampled images lose important high-frequency details, using them alone is suboptimal. However, we conjecture that integrating images at multiple resolutions can preserve high-frequency details, while boosting the importance of transformation-invariant signals. Therefore, we introduce a multi-resolution PRNU weighting strategy at inference time. Specifically, for a predefined set of $r$ resolutions $\{(h_i, w_i)\}_{i=1}^r$, we rescale the input and compute multiple similarity scores between the camera fingerprint $K$ and the image residual $R$. The similarity score is then obtained as a weighted combination of these values, as follows:
\begin{equation}
\label{score}
    s(K, R, r) = \sum_{i=1}^r \alpha^i \cdot E_\phi(K^i \odot R^i),
\end{equation}
where $K^i$ is the camera fingerprint at the $i$-th resolution, $R^i$ is the image noise residual at the $i$-th resolution and $\alpha^i = \frac{\max\left(h_i,w_i\right)}{\max_{j=1}^r\left(\max\left(h_j,w_j\right)\right)}$ represents the weight for the $i$-th resolution. Essentially, via Eq.~\eqref{score}, we normalize the resolutions into weights by dividing each resolution by the maximum resolution. As a result, the maximum resolution obtains a weight of 1, while all other resolutions are scaled proportionally relative to it. As noted earlier, this choice is motivated by the strong correlation between resolution and performance, observed in some preliminary experiments.

\noindent
\textbf{Joint prediction.}
We emphasize that the proposed neural-based comparison between the PRNU fingerprint and the PRNU noise residual is not intended to replace the traditional NCC-based comparison. In fact, the multi-resolution weighting scheme can also be applied to the NCC-based score. Thus, the two approaches are complementary and can even be combined to make the final prediction. We compute the sum of the neural-based score and the NCC-based score as the final output of the proposed pipeline, with each score individually incorporating the multi-resolution weighting, as follows:
\begin{equation}
\label{final_score}
    \begin{split}
        s_{\text{final}}(K, R, r)\!=\! \sum_{i=1}^r \alpha^i\!\cdot\!\left(E_\phi(K^i \!\odot\! R^i) \!+\! NCC(K^i, R^i)\right)\!.
    \end{split}
\end{equation}

\section{Experiments}

\noindent
\textbf{Overview.} 
For each camera, we estimate the PRNU fingerprint as described at the beginning of Section~\ref{method}. We then extract the noise residuals from all test images and compute their similarity to all fingerprints. The objectives of the experiments are: (i) to benchmark several methods on PRNU-Bench; (ii) to compare our approach with state-of-the-art methods; (iii) to assess the benefits of individual components of the proposed method via ablation studies.


\noindent
\textbf{Metrics.} We measure the performance by employing standard retrieval metrics, including the area under the receiver operating characteristic curve (AUC), the equal error rate (EER), and the top-1 and top-5 accuracy rates. 

\noindent
\textbf{Evaluation setup.} To increase the difficulty of the task, in Eq.~\eqref{prnu_estimate}, we restrict the number of images for PRNU fingerprint estimation to $n=5$ per camera. This setup reflects realistic conditions and makes the task more practical.

\noindent
\textbf{Additional dataset.}
In addition to conducting experiments on the newly introduced PRNU-Bench, we demonstrate the generalization capacity of our pipeline via experiments on the Dresden dataset~\cite{Thomas-SAC-2010}. This dataset includes 73 cameras with a total of 15,025 images. Following the same protocol as for PRNU-Bench, we estimate the PRNU fingerprint using only 5 images per device, while keeping 50 of the remaining images for evaluation, as in~\cite{cozzolino-JIS-2020}. We emphasize that we do not train our model on Dresden. Instead, we use the checkpoint trained on PRNU-Bench. 

\begin{table}[t]
    \centering
    \small{
    \begin{tabular}{ccccc}
    \toprule
         Training & \multirow{2}{*}{AUC $\uparrow$} & \multirow{2}{*}{EER $\downarrow$} & \multicolumn{2}{c}{Accuracy (\%) $\uparrow$}  \\
         Objective & & & Top-5 & Top-1\\
         \midrule
         
         Siamese loss & 0.6862 & 0.3729 & 11.90 & 2.61\\
         Triplet loss & 0.6726 & 0.3852 & 13.39 & 3.45 \\
         InfoNCE loss & 0.7205 & 0.3423 & 15.94 & 3.56 \\
         NCC-based baseline & 0.7849 & 0.4142 & 32.58 & 13.31\\
         BCE loss (ours) & \textbf{0.8588} & \textbf{0.2237} & \textbf{46.38} & \textbf{20.11} \\
         \bottomrule
    \end{tabular}
    }
    \vspace{-0.2cm}
    \caption{Results with various training objectives for the neural-based comparison. Our model based on binary cross-entropy is the only approach that outperforms the NCC-based baseline. The baseline does not employ any additional model beyond the denoiser. The image resolution is $768 \times 768$. Best scores are in bold. Arrows indicate when better scores correspond to higher ($\uparrow$) or lower ($\downarrow$) values.}
    \label{tab:ablation_training_method}
    \vspace{-0.2cm}
\end{table}

\noindent
\textbf{Baselines.} We compare against three baselines. One is a traditional denoising approach~\cite{Goljan-EI-2009} based on denoising filters. The other two employ denoising autoencoders, namely Restormer~\cite{Zamir-CVPR-2022} and DRUNet~\cite{Zhang-TPAMI-2021}, to extract noise residuals. 

\begin{table*}[t]
    \centering
    \small{
    \begin{tabular}{cccccc}
    \toprule
         \multirow{2}{*}{Dataset} & 
         \multirow{2}{*}{Method} & \multirow{2}{*}{AUC $\uparrow$} & \multirow{2}{*}{EER $\downarrow$} & \multicolumn{2}{c}{Accuracy (\%) $\uparrow$}  \\
         & & & & Top-5 & Top-1\\
         \midrule
         \multirow{4}{*}{PRNU-Bench (Ours)} & Classic \cite{Goljan-EI-2009} & 0.8713 & 0.2069 & 52.25 & 29.13 \\
         & DRUNet \cite{Zhang-TPAMI-2021} & 0.7678 & 0.3435 & 72.68 & 51.97 \\
         & Restormer \cite{Zamir-CVPR-2022} & 0.9092 & 0.1761 & 75.17 & 58.97\\
         &Ours & \textbf{0.9673} ({\color{ForestGreen}+0.0581}) & \textbf{0.0973} ({\color{ForestGreen}-0.0788}) & \textbf{86.14} ({\color{ForestGreen}+10.97}) & \textbf{73.65} ({\color{ForestGreen}+14.68})\\
                  \midrule

         \multirow{4}{*}{Dresden~\cite{Thomas-SAC-2010}} & Classic \cite{Goljan-EI-2009} & 0.8169 & 0.2576 & 69.00  & 44.31 \\
         & DRUNet \cite{Zhang-TPAMI-2021} & 0.7917 &  0.2903 & 83.31 & 64.70 \\
         & Restormer \cite{Zamir-CVPR-2022} & 0.8351 & 0.2556 & 82.23 & 65.85\\
         &Ours & \textbf{0.9012} ({\color{ForestGreen}+0.0661}) & \textbf{0.1700} ({\color{ForestGreen}-0.0856}) & \textbf{83.31} ({\color{ForestGreen}+1.08}) &  \textbf{68.69} ({\color{ForestGreen}+2.84})\\
         \bottomrule
    \end{tabular}
    }
    \vspace{-0.25cm}
    \caption{Results of our approach vs.~three baselines: classic method based on denoising filters~\cite{Goljan-EI-2009}, DRUNet~\cite{Zhang-TPAMI-2021} and Restormer~\cite{Zamir-CVPR-2022}. Our method brings substantial performance gains across all metrics. Best scores are in bold. Arrows indicate when better scores correspond to higher ($\uparrow$) or lower ($\downarrow$) values.}
    \label{tab:main_results}
    \vspace{-0.3cm}
\end{table*}

\noindent
\textbf{Hyperparameters.} 
To extract the PRNU signals from images, we alternatively integrate both Restormer~\cite{Zamir-CVPR-2022} and DRUNet~\cite{Zhang-TPAMI-2021} into our pipeline. Among them, we select Restormer, due to its superior performance as an individual baseline. 
We set the number of resolutions in Eq.~\eqref{final_score} to $r=2$, where the selected resolutions are $1024 \!\times\! 1024$ and $1400 \!\times\! 1400$. The images in PRNU-Bench are stored at their original resolution. To generate the previous resolutions, we first apply reflect padding to obtain an aspect ratio of $1\!:\!1$. Then, the images are resized using bicubic interpolation. Additional resolutions are investigated in an ablation study. The image encoder for our neural-based comparison is a ResNet-50 model~\cite{He-CVPR-2016}. We flatten the feature maps resulting from the last convolutional layer and attach a sigmoid layer for binary classification. We start the training of this model from a checkpoint pre-trained on ImageNet~\cite{Deng-CVPR-2009}. We set the number of training epochs to $50$, and we employ the Adam optimizer with a fixed learning rate of $10^{-3}$ and a mini-batch size of $8$.

\noindent
\textbf{Preliminary study. }
In preliminary experiments, we explore alternative training strategies for our image encoder, that are not based on Hadamard products. The goal of these experiments is to compare our Hadamard-based method with contrastive learning approaches. All the experiments are conducted in the same setup, at a resolution of $768 \times 768$ pixels, using Restormer for denoising.
During training, we leverage triplets composed of PRNU fingerprints, positive noise residuals, and negative noise residuals. 
Depending on the objective, these triplets are either treated as two separate pairs, \ie (fingerprint, positive noise) and (fingerprint, negative noise), or as full triplets. The comparative results are presented in Table~\ref{tab:ablation_training_method}. The first row in the table corresponds to the standard contrastive loss based on Siamese networks~\cite{Hadsell-CVPR-2006}, where inputs are embedded into a shared feature space and the objective is to minimize distances between positive pairs, while maximizing those between negative pairs. The second approach is based on triplet loss~\cite{Schroff-CVPR-2015}, in which the fingerprint plays the role of an anchor. The goal is to enforce the anchor-positive distance to be smaller than the anchor-negative distance, by at least a predefined margin. The third row corresponds to the InfoNCE loss~\cite{Chen-ICML-2020}, in which fingerprints are again used as anchors. The model is trained to maximize the similarity between an anchor and its positives, while minimizing similarity with all negatives. Overall, these contrastive learning approaches fail to surpass the NCC-based baseline, while performing significantly lower than the proposed Hadamard-based method. The results clearly indicate that the novel design of our training pipeline is superior.

\begin{table*}[t]
    \centering
    \small{
    \begin{tabular}{cccccc}
    \toprule
         Comparison &
         \multirow{2}{*}{Resolution} & \multirow{2}{*}{AUC $\uparrow$} & \multirow{2}{*}{EER $\downarrow$} & \multicolumn{2}{c}{Accuracy (\%) $\uparrow$}  \\
         Method & & & & Top-5 & Top-1\\
         \midrule
         NCC & $768 \times 768$ & 0.7849 & 0.4142 & 32.58 & 13.31 \\
         Ours & $768 \times 768$ & 0.8602 ({\color{ForestGreen}+0.0753}) & 0.2228 ({\color{ForestGreen}-0.1920}) & 46.81 ({\color{ForestGreen}+14.23}) & 20.41 ({\color{ForestGreen}+7.1})\\
         \midrule
         NCC & $1024 \times 1024$ & 0.8807 & 0.1964 & 65.04 & 40.97\\
         Ours &  $1024 \times 1024$ & 0.9320 ({\color{ForestGreen}+0.0513}) & 0.1428    ({\color{ForestGreen}-0.0536}) & 72.96 ({\color{ForestGreen}+7.92}) & 47.86 ({\color{ForestGreen}+6.89})\\
         \midrule
         NCC & $1400 \times 1400$ & 0.9092 & 0.1761 & 75.17 & 58.97 \\
         Ours & $1400 \times 1400$ & 0.9615 ({\color{ForestGreen}+0.0523}) & 0.1048 ({\color{ForestGreen}-0.0713}) & 84.64 ({\color{ForestGreen} +9.47}) & 70.83 ({\color{ForestGreen} +11.86})\\ 
         \bottomrule
    \end{tabular}
    }
    \vspace{-0.25cm}
    \caption{Results obtained with our neural-based method for comparing PRNU fingerprints and noise residuals show substantial performance gains, consistent across all tested resolutions. Arrows indicate when better scores correspond to higher ($\uparrow$) or lower ($\downarrow$) values.}
\label{tab:ablation_resolution_model_comparison}
\vspace{-0.3cm}
\end{table*}

\noindent
\textbf{Results on PRNU-Bench.} The main results are summarized in Table~\ref{tab:main_results}, where we report performance metrics for three NCC-based baselines (corresponding to the classic approach and the two denoisers) versus our method. 

\begin{figure}
    \centering
    \includegraphics[width=0.78\linewidth]{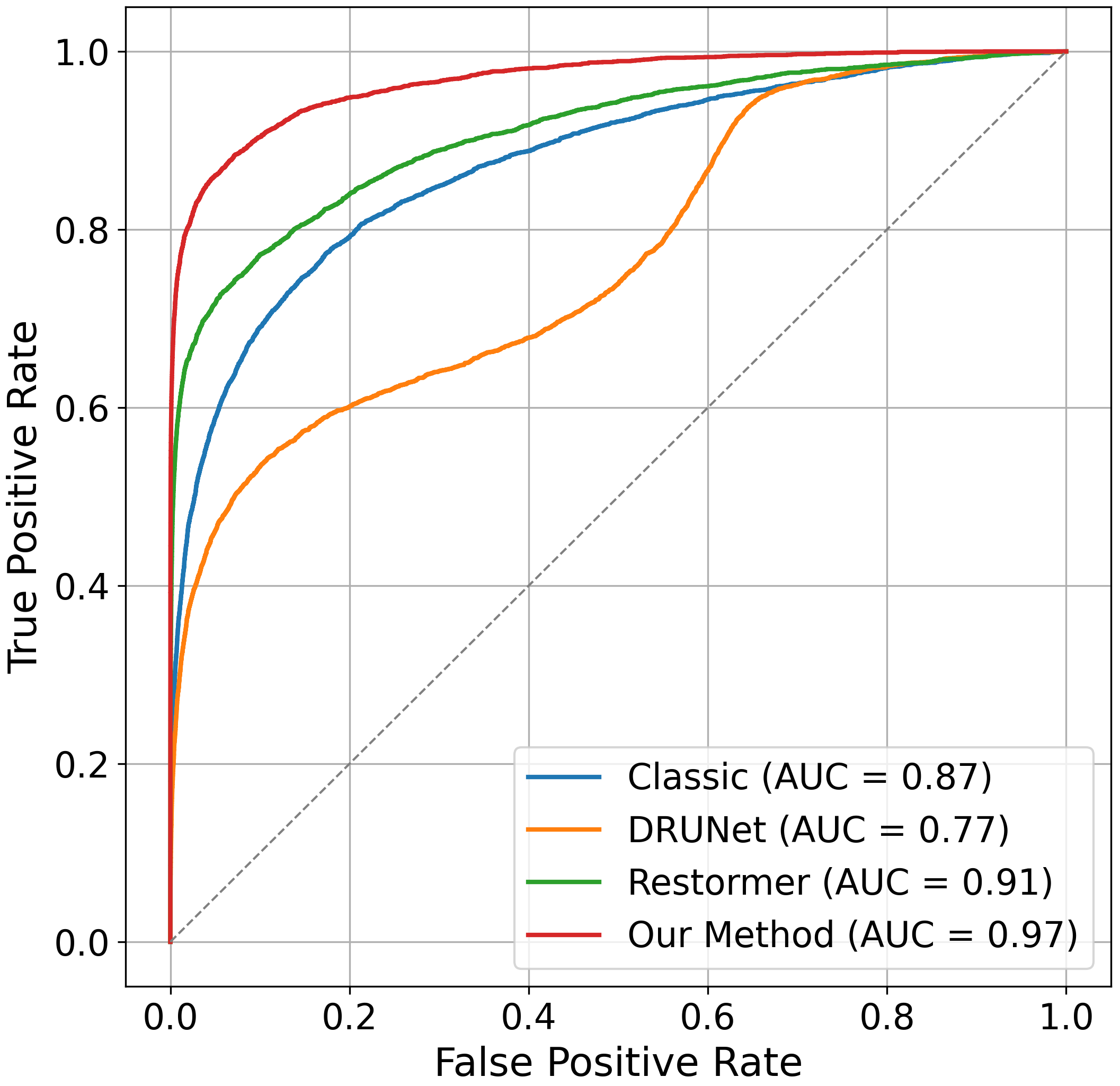}
    \vspace{-0.2cm}
    \caption{ROC curves our approach vs.~three baselines: classic denoising method, DRUNet, and Restormer. Best viewed in color.}
    \label{fig:ROC}
    \vspace{-0.4cm}
\end{figure}

The results show that the PRNU-Bench is a challenging dataset, since all three baselines achieve top-1 accuracy rates below $60\%$. Furthermore, the results clearly demonstrate that our approach outperforms the baselines in source camera identification, achieving substantially higher scores across all evaluation metrics. The ROC curves illustrated in Figure~\ref{fig:ROC} further confirm that our method achieves a significant performance improvement compared with the baseline methods. 

\begin{figure}
    \centering
    \includegraphics[width=0.95\linewidth]{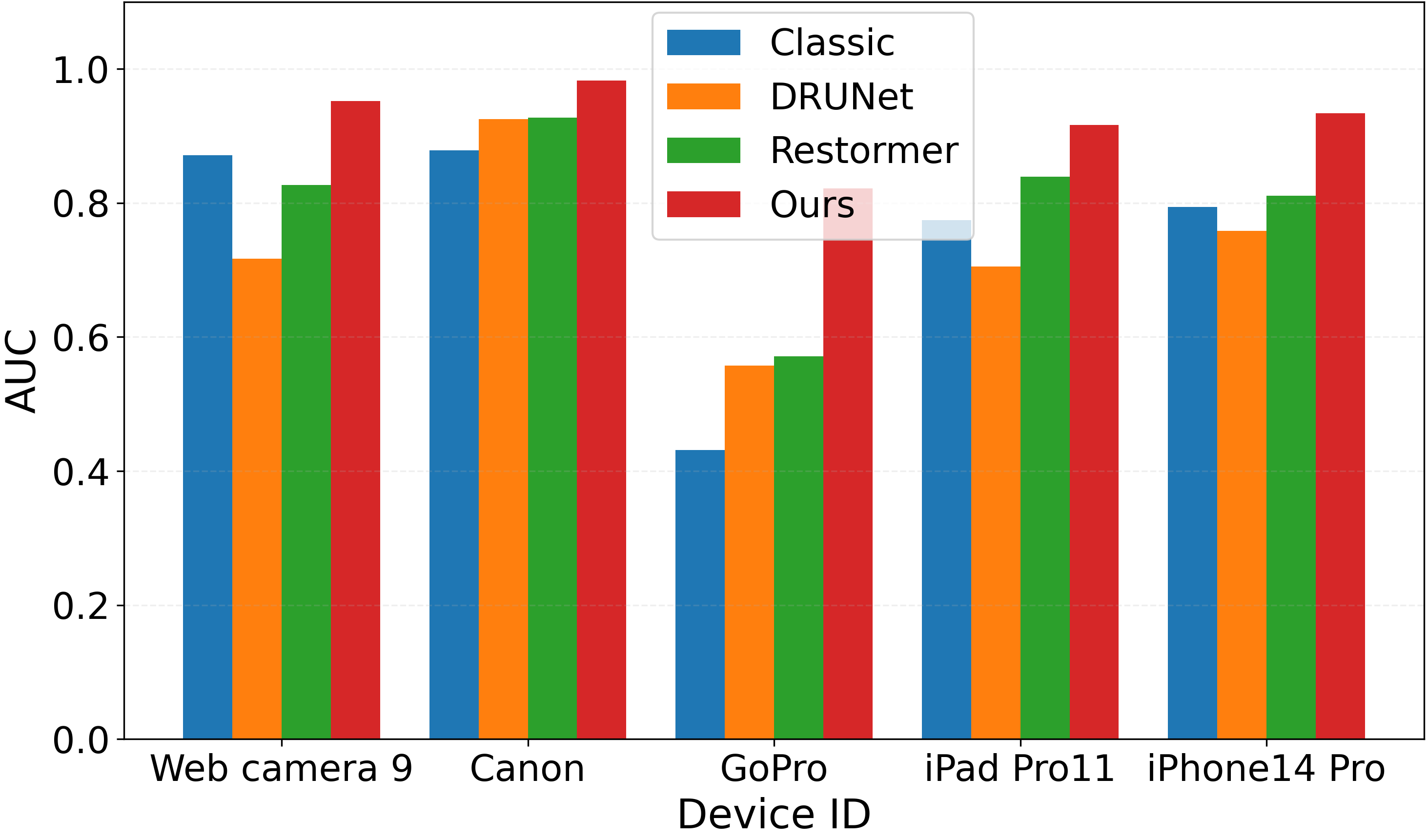}
    \vspace{-0.25cm}
    \caption{Per device AUC values of our approach vs.~the three baselines. There is one representative device from each type.}
    \label{fig:barplot}
    \vspace{-0.4cm}
\end{figure}

In Figure~\ref{fig:barplot}, we exhibit the AUC scores for a set of randomly chosen devices, taking one device from each device type included in the dataset. The selected cases demonstrate that our method is consistently better across various device types. Notably, when the baselines perform very close to random guessing (AUC $\approx\!0.5$), our method shows substantial improvements, \eg~as observed for the GoPro camera.

\noindent
\textbf{Results on Dresden.}
The results on the Dresden dataset~\cite{Thomas-SAC-2010} further demonstrate the generalization ability of our approach. Across all evaluation metrics, our method consistently outperforms the baselines. In particular, we observe an increase of more than 2\% in top-1 accuracy, underscoring the robustness of our model. Even larger relative gains are achieved for AUC and EER, where the improvements over the baselines are more substantial when compared with those obtained on PRNU-Bench.

\noindent
\textbf{Ablation studies.} Our method combines several components into a joint pipeline. To better understand the role and effectiveness of these components, we carry out a series of ablation studies. These ablation experiments not only allow us to isolate and quantify the importance of each component, but also to examine different configurations within each component. Through these empirical analyses, we aim to provide a better picture of how the proposed method achieves its improvements, and which contributions are the most important for the observed performance gains.

\begin{table}[t]
    \centering
    \setlength\tabcolsep{0.18em}
    \small{
    \begin{tabular}{cccccc}
    \toprule
         \multirow{2}{*}{Resolutions} & Aggregation & \multirow{2}{*}{AUC $\uparrow$} & \multirow{2}{*}{EER $\downarrow$} & \multicolumn{2}{c}{Accuracy (\%) $\uparrow$}  \\
           & Method & & & Top-5 & Top-1\\
         \midrule
         1400 & - & 0.9092 & 0.1761 & 75.17 & 58.97 \\
         \midrule
         768, 1024, 1400 & Mean & 0.9112 & 0.1625 & 75.50& 57.20\\
         1024, 1400 & Mean & 0.9131 &  0.1628 & 76.69 & 59.55\\
         \midrule
         768, 1024, 1400 & Eq.~\eqref{score} & \textbf{0.9141 } & 0.1606 & 76.91 &59.34 \\
         1024, 1400 & Eq.~\eqref{score} & \textbf{0.9141 }& \textbf{0.1602} &\textbf{77.06} & \textbf{60.16}\\
   \bottomrule
    \end{tabular}
    }
    \vspace{-0.25cm}
    \caption{Comparison of different resolution combinations and aggregation schemes. Best scores are in bold. Arrows indicate when better scores correspond to higher ($\uparrow$) or lower ($\downarrow$) values.}
\label{tab:resolution_aggregation}
\vspace{-0.3cm}
\end{table}

In Table~\ref{tab:ablation_resolution_model_comparison}, we present a comparison between the NCC-based similarity and our model-based similarity across multiple resolutions. We underline that in these experiments, we did not use Eq.~\eqref{final_score} to compute the similarity score. Instead, we only leverage the output of the model, without any weighting or combination with NCC. This setup ensures that the improvements observed in the ablation study can be attributed exclusively to the discriminative capability of the trained model itself, rather than to complementary mechanisms. The results clearly indicate that our neural-based approach consistently outperforms NCC, at all tested resolutions. This demonstrates not only the robustness of the proposed method to resolution changes, but also the key role that the neural-based similarity plays in driving the improvements observed in the final results.

In Table~\ref{tab:resolution_aggregation}, we present an ablation study evaluating the importance of the number of resolutions used in the similarity score computation, along with different strategies for aggregating these scores. The last two rows employ the weighting scheme presented in Eq.~\eqref{score}. First, we observe that simply averaging the scores is suboptimal and the performance is boosted when the resolution-based weights are employed. Second, when we include lower resolutions the performance is negatively impacted, this indicates that PRNU is significantly altered at these lower resolutions.

Lastly, in Table~\ref{tab:pipeline_components}, we report an ablation study that quantifies the contribution of each individual component of our pipeline. As discussed earlier, the principal driver of performance is the neural-based comparison of PRNU fingerprints and noise residuals, which provides the most substantial improvements over the baseline. Nevertheless, the results also highlight that the remaining components are far from negligible. Harnessing multi-resolution score aggregation and the complementary NCC-based comparison leads to consistent gains, which further boost the overall performance.

\begin{table}[t]
    \centering
    \setlength\tabcolsep{0.38em}
    \small{
    \begin{tabular}{ccccccc}
    \toprule
        \multicolumn{2}{c}{Comparison} & Multi- & \multirow{2}{*}{AUC $\uparrow$}& \multirow{2}{*}{EER $\downarrow$} & \multicolumn{2}{c}{Accuracy (\%) $\uparrow$} \\
         NCC & Model & Resolution & & & Top-5 & Top-1\\
    \midrule
    {\color{ForestGreen}\cmark}&{\color{Red}\xmark}& {\color{Red}\xmark}& 0.9092& 0.1761 & 75.17 & 58.97\\
    {\color{Red}\xmark}&{\color{ForestGreen}\cmark}& {\color{Red}\xmark}& 0.9615& 0.1051 & 84.67 & 70.75\\
    {\color{ForestGreen}\cmark}&{\color{Red}\xmark}& {\color{ForestGreen}\cmark}& 0.9141& 0.1602 & 77.06 & 60.16\\
    {\color{ForestGreen}\cmark}&{\color{ForestGreen}\cmark}& {\color{Red}\xmark}&0.9615& 0.1048 & 84.64 & 70.83\\
    {\color{Red}\xmark}&{\color{ForestGreen}\cmark}& {\color{ForestGreen}\cmark}&0.9636& 0.1025 & 85.83 & 72.41\\
    {\color{ForestGreen}\cmark}&{\color{ForestGreen}\cmark}& {\color{ForestGreen}\cmark}&\textbf{0.9673}& \textbf{0.0973} & \textbf{86.14} & \textbf{73.65}\\
    \bottomrule
    \end{tabular}
    }
    \vspace{-0.25cm}
    \caption{Ablation study of the main components of our pipeline. The largest improvement comes from our model-based comparison of PRNU fingerprints and noise residuals. This is further enhanced by combining scores across multiple resolutions and by additionally incorporating the NCC-based comparison. Best scores are in bold. Arrows indicate when better scores correspond to higher ($\uparrow$) or lower ($\downarrow$) values.}
    \label{tab:pipeline_components}
    \vspace{-0.3cm}
\end{table}

\section{Conclusion and Future Work}
In this work, we introduced PRNU-Bench, a large-scale dataset specifically designed to advance research in source camera identification. By including a diverse set of recent and unique cameras, PRNU-Bench offers a realistic and challenging evaluation environment. We also proposed a novel method that computes similarity scores by analyzing the Hadamard product between PRNU fingerprints and noise residuals, and demonstrated that incorporating multiple resolutions provides consistent performance improvements. Together, the dataset and the proposed methodology establish a solid foundation for future work in this area. We believe that subsequent research will build upon this benchmark to further improve accuracy and robustness, ultimately enhancing the practicality of source camera identification in real-world scenarios.

\vspace{0.1cm}
\noindent
\textbf{Acknowledgments.}
This work was supported by a grant of the Ministry of Research, Innovation and Digitization, CCCDI - UEFISCDI, project number PN-IV-P6-6.3-SOL-2024-2-0227, within PNCDI IV.

{
    \small
    \bibliographystyle{ieeenat_fullname}
    \bibliography{main}
}

\end{document}